\documentclass[runningheads]{llncs}

\usepackage{eccv}

\usepackage{eccvabbrv}

\usepackage{graphicx}
\usepackage{booktabs}
\usepackage{xcolor,graphicx}
\usepackage{tikz}
\usepackage{lipsum} 
\usepackage{float}
\usepackage{multirow}
\usepackage[accsupp]{axessibility}  

\usepackage{orcidlink}
\usepackage{xcolor}
\usepackage{caption}
\hypersetup{ colorlinks=true, linkcolor=black, citecolor=black, filecolor=black, urlcolor=blue }

\begin{document}
\title{XDen-1K: A Density Field Dataset of Real-World Objects}

\author{
Jingxuan Zhang\inst{1,2}$^{*}$ \and
Tianqi Yu\inst{1}$^{*}$ \and
Yatu Zhang\inst{1,2}$^{*}$ \and
Jinze Wu\inst{1} \and
Kaixin Yao\inst{1,2} \and
Jingyang Liu\inst{1} \and
Yuyao Zhang\inst{1}$^{\dagger}$ \and
Jiayuan Gu\inst{1}$^{\dagger}$ \and
Jingyi Yu\inst{1}$^{\dagger}$
}

\authorrunning{J.~Zhang et al.}

\institute{
ShanghaiTech University\\
\and
Deemos Technology
}

\maketitle
\vspace{-1em}
{\centering\small
\texttt{\{zhangjx12023,zhangyy8,gujy1,yujingyi\}@shanghaitech.edu.cn}\par
$^{*}$ Equal contribution.\quad
$^{\dagger}$ Corresponding authors.\par
}

  \begin{center}
  \includegraphics[width=0.99\textwidth]{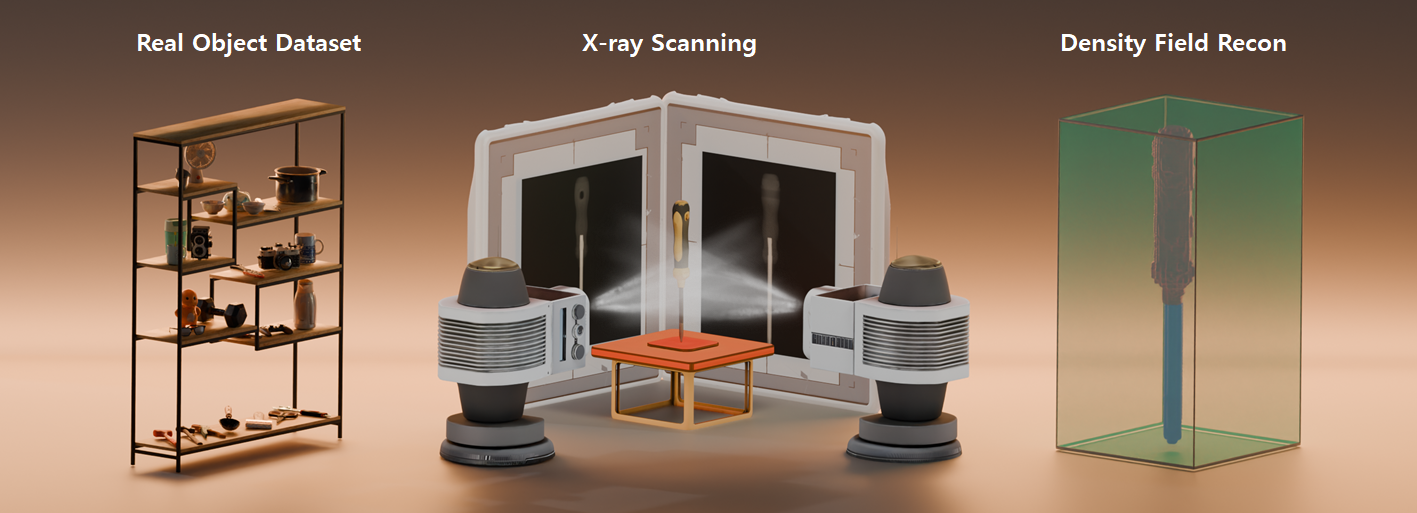}
  \captionof{figure}{We present XDen-1K, a multi-modal dataset of real-world objects, featuring paired biplanar X-ray scans and reconstructed density field.}
  
  \label{fig:teaser}
  \end{center}
\begin{abstract}
A deep understanding of the physical world is essential for robotic manipulation and physically realistic simulation. While current methods, including VLM-based and other learning-based approaches, have shown promise in physical property inference, their evaluation is often hindered by the lack of physically grounded reference data. To address this gap, we introduce XDen-1K, the first large-scale multimodal dataset that provides physically grounded density field for real-world objects. XDen-1K comprises 1,000 real-world objects spanning 137 categories, with comprehensive data for each object, including a high-resolution, carefully curated 3D geometric model with part-level annotations and paired real-world biplanar X-ray scans. In addition, XDen-1K includes high-fidelity volumetric density field reconstructed from sparse biplanar X-ray views via a novel optimization framework. XDen-1K also provides a benchmark for density estimation  and enables X-ray-conditioned volumetric segmentation. Experiments further demonstrate that the center-of-mass prior derived by XDen-1K can improve robotic manipulation performance. By providing real-world X-ray scans and physics-consistent density field, XDen-1K establishes a foundation for advancing physical property inference and embodied AI. Additional details and dataset resources are available at \url{https://xden-1k.github.io/} and \url{https://huggingface.co/datasets/zhangjxx/XDen-1K}.

\end{abstract}

\section{Introduction}
\label{sec:intro}

Objects are more than their appearance: beyond color and texture, they possess internal composition — materials, thickness, voids, reinforcements, and mass distribution — that determine how they function, fail, and can be used. Modeling these physical attributes, coupled with visual inferences of shape and texture, enables more reliable physical reasoning~\cite{CLEVRER2020ICLR,bakhtin2019phyre,Li2024IPHYRE}, physically-based simulations~\cite{Chang2025Winding,kumar2019scalablemodularmaterialpoint}, and, most recently, object and scene reconstructions~\cite{yao2025castcomponentaligned3dscene,xiang2024structured}. For robots, these ``hidden'' properties are particularly useful for inferring where mass is concentrated~\cite{slota2012stability, Xu2024GaussianPropertyIP} (center of mass, moments of inertia), how stiff or compliant parts are, what forces and grasps are safe and effective, how objects will deform or fracture, which tools to use, etc.

Humans infer physical properties by combining visual cues (multi-view geometry, symmetry, material priors) and interactions (e.g. sound, deformation, context)~\cite{wang2025vggtvisualgeometrygrounded, NEURIPS2024_d7af02c8, tung2023physionevaluatingphysicalscene, jiang2025phystwinphysicsinformedreconstructionsimulation}. At surface level, recent success in material-property estimation~\cite{10.1145/3618358, huang2024materialanythinggeneratingmaterials} (e.g. SVBRDF/BSDF, albedo/roughness, microgeometry) is largely attributed to the availability of controlled, large-scale visual datasets. Density, in contrast, is inherently volumetric and, thereby, ``invisible'' to regular RGB cameras. The latest efforts~\cite{Cao2025PhysX3DP3, Xu2024GaussianPropertyIP, ahmed20253dcompat200languagegroundedcompositionalunderstanding} have focused on labeling parts by associating their appearance with material types. Resulting datasets are therefore synthetic and lack real-world validation. A real-world density dataset will facilitate learning in physically meaningful units and expose models to realistic multi-material mixtures and manufacturing artifacts. It also provides the urgently needed common benchmarks for current and future visual inference tasks.

In this paper, we present \textbf{XDen-1K}: an X-ray-imaging-based \textbf{density field dataset} of \textbf{1,000} real-world objects. XDen-1K is the first of its type, with objects covering sizes from small hardware (e.g. screw, flashlight, key) to medium household goods (e.g. kettle, bicycle helmet, power tool at $\sim20-30$ cm), and to larger items (e.g. cookware, sports equipment, furniture). The most direct brute-force approach to acquiring a ground-truth density field is computed tomography (CT), which was initially developed for medical and industrial inspection. However, high-resolution CT is time-consuming and costly—typical scans take 15–60 minutes and can cost \$100--\$500 per object, so 1,000 objects imply hundreds to thousands of scanner-hours and a six-figure budget. Even with CT, the so-called beam hardening effect will make recovering internal information more difficult.

Density field reconstruction begins with biplanar X-ray imaging of each object and the construction of its 3D mesh. To constrain this highly underdetermined problem, manually annotated object parts are used as structural priors, with each part assumed to be composed of a homogeneous material. Based on Beer-Lambert law, a differentiable optimization framework is introduced to match rendered X-ray projections with real scans and optimize the linear attenuation coefficient (LAC) of each part, which is then converted to density. Following this pipeline, approximate density field are recovered for 1,000 objects, together with their meshes, part annotations, and biplanar X-ray scans. To further validate the reconstruction, a small-scale CT dataset is collected for evaluation.

Based on XDen-1K, our paired X-ray and density field dataset, we introduce three subtasks: benchmarking physical reasoning methods~\cite{Le2025PixieFA,Zhai2024PhysicalPU,Xu2024GaussianPropertyIP},  X-ray-conditioned volumetric segmentation and CoM-aware robot manipulation. These three tasks take full advantage of our data, where X-ray provides strong structural and material prior and the density field provides physical information on both static (mass, center of mass) and dynamic (moment of inertia) properties of an object. We commit to \textbf{releasing our dataset publicly} to facilitate multi-modal learning and physical reasoning.

\vspace{0.5em}
\textbf{Our contributions:}

\vspace{0.5em}
1. We propose the \textbf{XDen-1K} dataset, the first large-scale real-world object dataset, coupled with both \textbf{biplanar X-ray images} and \textbf{density fields}, which serves as a benchmark on related physical reasoning tasks.

2. We develop a differentiable optimization framework that takes part-segmented object geometry and real biplanar X-ray scans as inputs to reconstruct each object's volumetric density field.

3. We demonstrate the effectiveness of  X-ray imaging as an additional modality on 3D volumetric segmentation task.

4. We show that center-of-mass priors derived from XDen-1K’s density information can improve robotic grasping performance.
\section{Related Work}
\label{sec:Related Work}

\paragraph{Density Inference and Physics-Aware Modeling.}

A broad line of work seeks to endow 3D vision with physical awareness by predicting material or density-related properties directly from visual input~\cite{Li2023PACNeRFPA,Xie2025Vid2SimRA,huang2024dreamphysics,Zhai2024PhysicalPU,Xu2024GaussianPropertyIP,Le2025PixieFA,cao2025physxanythingsimulationreadyphysical3d,shuai2025pugs}. Optimization-based approaches combine differentiable physics with neural scene representations to fit elastic or material parameters from observed motion, but they are typically evaluated on synthetic or tightly controlled setups, making it unclear how well they transfer to real objects. More recent feed-forward methods instead regress spatially varying material fields from learned 3D features. PIXIE~\cite{Le2025PixieFA}, for example, learns a supervised mapping from distilled feature volumes to per-voxel material parameters, while GaussianProperty~\cite{Xu2024GaussianPropertyIP} and PuGS\cite{shuai2025pugs} assigns physical attributes to 3D Gaussians using segmentation and vision-language priors. Collectively, these methods demonstrate that appearance encodes rich physical cues, but in the absence of real volumetric ground truth, their predictions remain difficult to rigorously validate. Works like PhysX-3D~\cite{Cao2025PhysX3DP3} and PhysX-Anything~\cite{cao2025physxanythingsimulationreadyphysical3d} aim to build physics-ready assets for simulation and also augment large 3D model collections with absolute scale, material categories, affordances, and kinematic attributes. Datasets like 3DCoMPaT200~\cite{ahmed20253dcompat200languagegroundedcompositionalunderstanding} contain large amounts of 3D assets and their material composition. 

These resources mark an important step toward learning physically grounded representations; yet, they are constructed from synthetic  3D assets and rely on assumed or inferred physical properties rather than measurements. None provide large-scale, real-world volumetric density fields against which both prediction methods and physics-aware generative models can be quantitatively evaluated. In contrast, our XDen-1K dataset is built from calibrated biplanar X-ray measurements,  offering empirically grounded density fields for 1,000 real objects and furnishing the missing real-world benchmark for volumetric density estimation.

\paragraph{X-ray and CT imaging.}
X-rays are high-energy electromagnetic waves characterized by high photon energy and a low probability of interaction with matter.
Unlike infrared or visible light, which are easily absorbed or reflected at the surface, 
X-rays can penetrate most materials and reveal internal structures without causing damage~\cite{glover2016objectively}. This strong penetration capability enables the observation and reconstruction of 3D material distributions from 2D projections. Computed Tomography (CT), as one of the most accurate 3D reconstruction techniques, acquires dense angular projections—collectively referred to as a sinogram—by scanning an object from multiple viewpoints. Classical reconstruction algorithms~\cite{feldkamp1984practical,rodet2004cone,lin2014efficient, nuyts2013modelling} can effectively recover volumetric information from these dense sinograms. Consequently, CT reconstruction plays a crucial role in medical diagnosis, industrial defect detection, and scientific imaging~\cite{sun2022review,ginat2014advances}.

However, a CT scan often incurs high acquisition cost and long scanning time, while traditional algorithms suffer from severe streak and beam-hardening artifacts in the presence of high-density materials. Considerable research has therefore focused on reducing CT scan time via sparse-view CT reconstruction, which lowers the number of projection angles required for reconstruction. For example, SAX-NeRF~\cite{cai2024structure} and SCOPE~\cite{wu2023self} use continuum modeling based on Neural Radiance Fields (NeRF)~\cite{mildenhall2020nerfrepresentingscenesneural}, where sparse angular projections are constrained and fused with viewpoints in an implicit space, thereby filling in missing viewpoint information in the learned dense volume representation. Nevertheless, at least dozens of projections are typically required to achieve acceptable quality; otherwise, strong sparse artifacts remain inevitable.

To further reduce acquisition time and radiation dose, biplanar X-ray imaging has emerged as a promising alternative~\cite{melhem2016eos}. By simultaneously projecting from two orthogonal viewpoints, this technique partially compensates for the loss of depth information inherent in single-view X-rays. Recent advances have demonstrated its potential in clinical diagnosis~\cite{chen2024automatic,berg2020experiences}, providing richer structural cues and improved diagnostic accuracy compared to single-view X-ray imaging.
Several studies have explored reconstructing 3D CT-like volumes from biplanar X-rays using either generative models~\cite{ying2019x2ct,jeong2025dx2ct} or regression-based mapping networks~\cite{huang2024generalizable,kyung2023perspective}. These approaches highlight the feasibility of biplanar imaging as a low-cost, low-time, and low-dose substitute for traditional CT. However, they typically rely on large paired datasets of high-quality CT and corresponding X-rays for supervision. For non-medical objects where such paired data are unavailable, how to effectively incorporate prior knowledge and achieve faithful 3D reconstruction remains a challenging open problem.

\paragraph{Part-Level Understanding.}
Decomposing fused 3D reconstructions into semantic parts is still far from solved, largely because reliable part boundaries require reasoning about global 3D geometry rather than local appearance cues alone. A common strategy is therefore to borrow strength from the well-established 2D segmentation methods and project or fuse these predictions back onto the surface~\cite{zhong2024meshsegmenter,ma20243d},but multi-view fusion often introduces cross-view inconsistencies and uneven granularity. Alternatively, 3D-native approaches learn part-sensitive representations directly. Such methods tend to yield more stable and coherent segmentations on  surfaces~\cite{ma2025p3sam,Liu2025PARTFIELDL3,yang2024sampart3d}, and some go a step further by explicitly generating part decompositions ~\cite{Zhang2025BANGD3,yang2025omnipart}. Moreover,part segmentation is not restricted to mesh surfaces: it has been actively explored in  point cloud representation~\cite{qi2017pointnet,zhao2021pointtransformer,zhou2025pointsam}, 3D Gaussian representations~\cite{cen2023saga,10658260,gaussian_grouping}, and radiance-field-based models~\cite{cen2023segment,garfield2024}. Most current method takes RGB cue as the only guidance,which can be ambiguous under texture variation. In this work, we introduce X-ray as an extra condition to guide the segmentation process, yielding more consistent result. Furthermore, the X-ray dataset we provide serves as a valuable benchmark for future multimodal-guided segmentation tasks.

\label{sec:dataset}
\begin{figure}[t]
    \centering
    \includegraphics[width=1\linewidth]{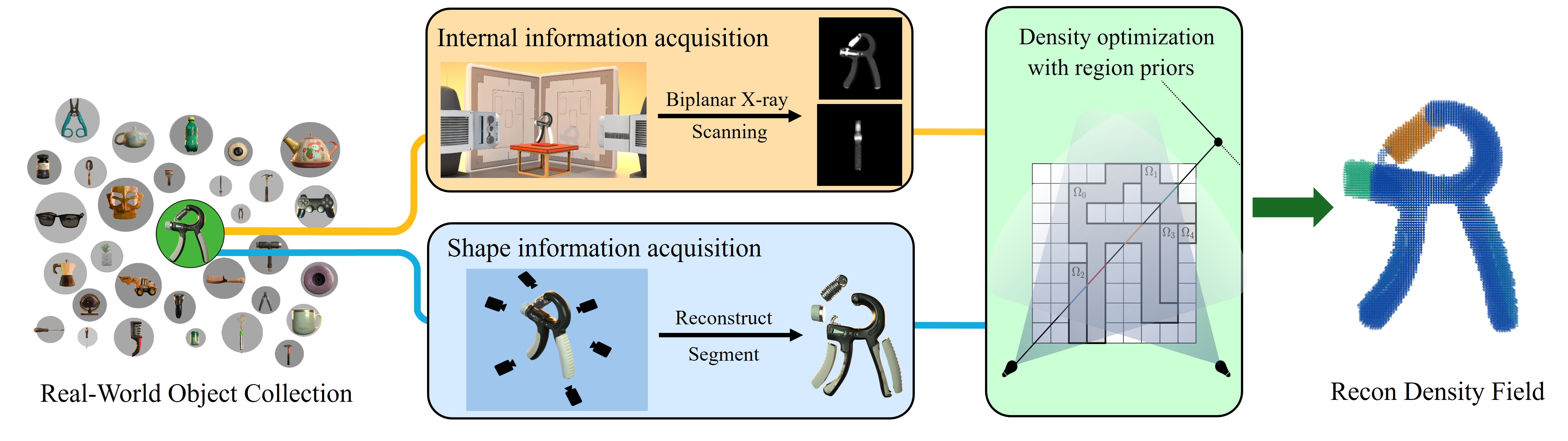}
    \caption{\textbf{Overview of XDen-1K generation pipeline.} Multi-view RGB images are used to reconstruct a surface mesh, and manually labeled segments provide accurate part-level shape information. Orthogonal biplanar X-rays capture internal attenuation signals. The recovered shape priors are then incorporated into an X-ray optimization process to estimate the volumetric density field.}
    \label{fig:pipeline}
    \vspace{-1em}
\end{figure}

\section{Density field reconstruction from biplanar X-ray images}
\subsection{Preliminary: X-ray Imaging}
\label{sec:x-ray}

X-rays possess the ability to penetrate matter. When an X-ray beam passes through an object, its intensity attenuates exponentially according to the Beer-Lambert law. For a ray $\mathbf{r}$ traversing a volumetric object, the detected intensity $I$ is related to the incident intensity $I_0$ as
\begin{equation}
I = I_0 e^{-\int_{L} \mu(x) \mathrm{d}x},
\label{eq:foward_model}
\end{equation}
where $\mu(x)$ denotes the linear attenuation coefficient (LAC) at spatial location $x$, which characterizes the probability per unit length that an X-ray photon is absorbed or scattered by the material.
Taking the logarithm of both sides yields the projection measurement $p$:
\begin{equation}
p = -\ln\left(\frac{I}{I_0}\right) = \int_{L} \mu(x) \mathrm{d}x,
\label{eq:projection_integral}
\end{equation}
which corresponds to the line integral of the LAC along the ray trajectory. In practice, the detector measures the transmitted intensity $I$, and the reference air-scan intensity $I_0$ is pre-calibrated. The logarithmic transformation converts the exponential attenuation into a linear integral form, establishing the foundation of tomographic reconstruction.
The LAC value $\mu(x)$ serves as a quantitative descriptor of material composition and density.
\begin{figure}[t]
    \centering
    \includegraphics[width=0.8\linewidth]{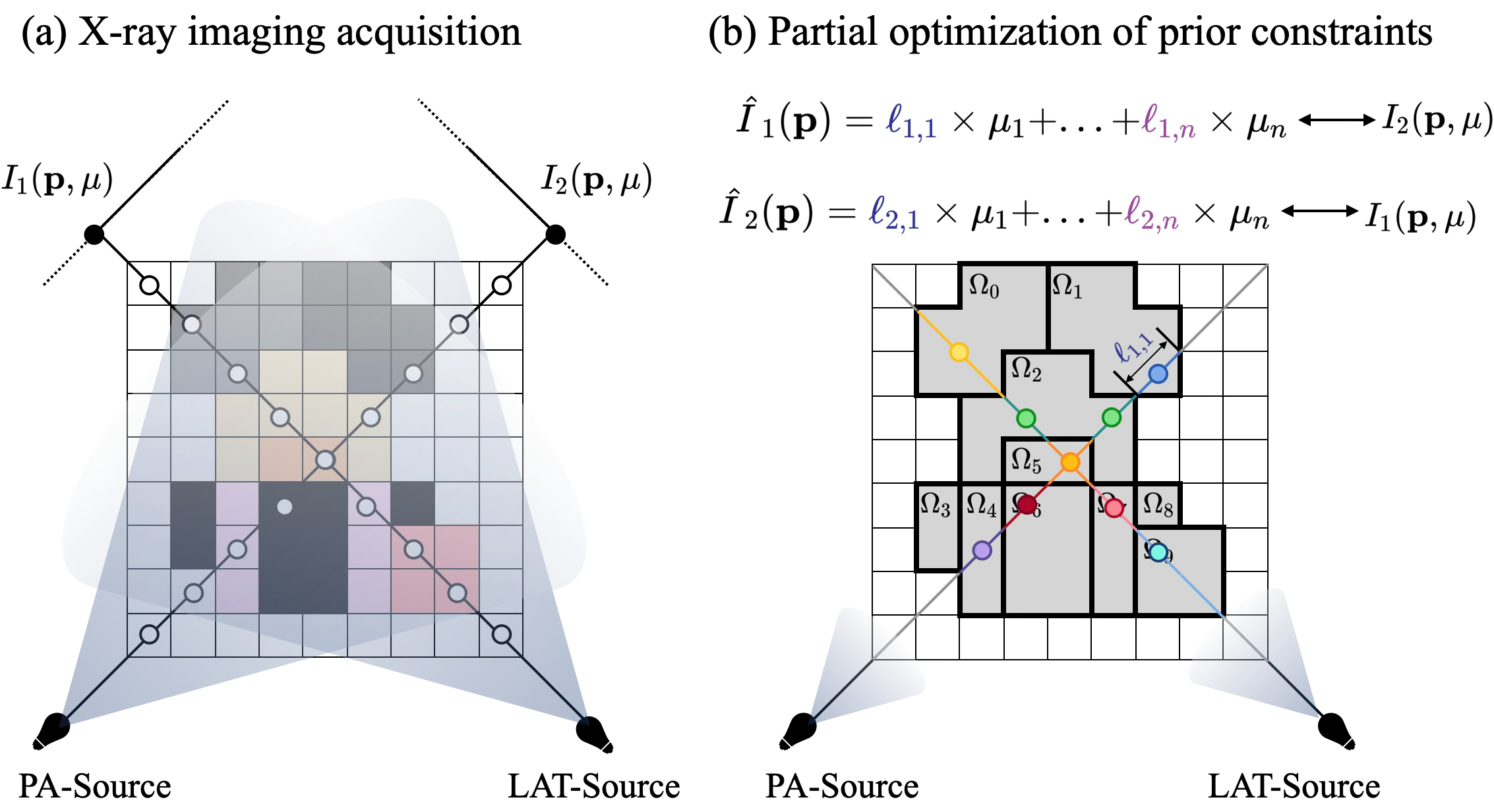}
    \caption{\textbf{Differentiable rendering optimization.} Region-wise attenuation coefficients are optimized to match captured projections, producing a 3D attenuation field.}
    \vspace{-1em}
    \label{fig:recon}
\end{figure}
Physically, the LAC depends on both the density $\rho$, and the mass attenuation coefficient (MAC) $\frac{\mu}{\rho}(E)$, which is a material-dependent quantity determined by atomic composition and photon energy $E$:
\begin{equation}
\mu(E) = \rho \cdot \frac{\mu}{\rho}(E).
\label{eq:lac_density_relation}
\end{equation}
According to reference database XCOM~\cite{berger1987xcom} from the National Institute of Standards and Technology (NIST), the MACs of most common materials (e.g., water, polymers, biological tissues) converge to approximately $0.17~\mathrm{cm^2/g}$ at a certain energy level (e.g., $100$~keV), as detailed in \cref{tab:material_properties}. Consequently, for such materials, $\mu$ exhibits an approximately linear relationship with density, enabling direct density estimation from reconstructed LAC values:
\begin{equation}
\rho = \frac{\mu}{(\mu/\rho)} = \frac{\mu}{0.17}.
\label{eq:estimate_rho}
\end{equation}
This principle underpins quantitative X-ray imaging, allowing reconstructed attenuation maps to be interpreted as physically meaningful density fields.

\begin{table}[t]
\centering
\caption{Mass attenuation coefficient (MAC), density, and linear attenuation coefficient (LAC) of some representative materials at $100$ keV ray energy.}
\label{tab:material_properties}
\small
\setlength{\tabcolsep}{12pt}
\renewcommand{\arraystretch}{1.0}

\begin{tabular}{@{}l l c c c@{}}
\toprule
\textbf{Material} &
\shortstack{\textbf{Chemical}\\\textbf{Formula}} &
\shortstack{\textbf{MAC}\\(cm$^2$/g)} &
\shortstack{\textbf{Density}\\(g/cm$^3$)} &
\shortstack{\textbf{LAC}\\(cm$^{-1}$)} \\
\midrule
Water       & H$_2$O              & 0.17 & 1.00   & 0.17 \\
Air         & N$_2$, O$_2$         & 0.15 & 0.0012 & $1.8\times 10^{-4}$ \\
Glass       & SiO$_2$              & 0.18 & 2.40   & 0.432 \\
Plastic (PP)  & (C$_3$H$_6$)$_n$    & 0.18 & 0.90   & 0.16 \\
Rubber      & (C$_5$H$_8$)$_n$     & 0.19 & 1.20   & 0.22 \\
Wood        & C$_6$H$_{10}$O$_5$   & 0.16 & 0.80   & 0.12 \\
Aluminum    & Al                  & 0.19 & 2.70   & 0.51 \\
\bottomrule
\end{tabular}
\vspace{-1em}
\end{table}

\subsection{Reconstruction Algorithms}
\label{sec:recon}
The \textbf{reconstruction objective} is to derive density field over the target object. Inverting Eq.~\eqref{eq:projection_integral} over all rays can recover the 3D spatial distribution of $\mu(x)$, the LAC field and by 
Eq.~\eqref{eq:lac_density_relation},the reconstructed LAC field can be converted to density field.
While  classical algorithms such as Filtered Backprojection (FBP) in CT reconstruction provide accurate 3D reconstructions, its high cost and long acquisition times make it prohibitive for large-scale datasets.  
In contrast, single-plane X-ray imaging is faster and more economical, yet it completely loses depth information by integrating attenuation coefficients along the projection direction.  
Consequently, recovering a 3D density field from a single X-ray projection is a severely ill-posed inverse problem.  
Therefore, we adopt a pragmatic approach that utilizes \textbf{two orthogonal, biplanar X-ray projections}, regularized by a strong geometric prior.  
This strategy leverages the complementary depth cues from the two projections and constrains the solution space through known geometry, thereby making the reconstruction problem tractable.

The process of obtaining the estimated density field is shown in \cref{fig:recon}. Given a captured ground-truth biplanar X-ray images $I_i$ and a well-structured segmentation that partitions the image into $K$ disjoint regions $\{\Omega_k\}_{k=1}^{K}$, each region corresponds to a homogeneous material with a constant linear attenuation coefficient (LAC) $\mu_k$.
The volumetric attenuation field $\mu(\mathbf{x})$ is thus modeled as a piecewise-constant function:
\begin{equation}
\mu(\mathbf{x}) = \sum_{k=1}^{K} \mu_k \, \mathbf{1}_{\Omega_k}(\mathbf{x}),
\label{eq:region_mu}
\end{equation}
where $\mathbf{1}_{\Omega_k}(\mathbf{x})$ is the indicator function:
\[
\mathbf{1}_{\Omega_k}(\mathbf{x}) =
\begin{cases}
1, & \text{if } \mathbf{x} \in \Omega_k,\\
0, & \text{otherwise.}
\end{cases}
\]

Let $\mathcal{L}_i(\mathbf{p})$ denote the ray of biplanar X-ray acquisition geometry corresponding to pixel $\mathbf{p}$ on detector plane $i \in \{1, 2\}$.  
The measured intensity $I_i(\mathbf{p})$ at pixel $\mathbf{p}$ follows \cref{eq:projection_integral}.

\begin{figure}[t]
    \centering
    \includegraphics[width=1\linewidth]{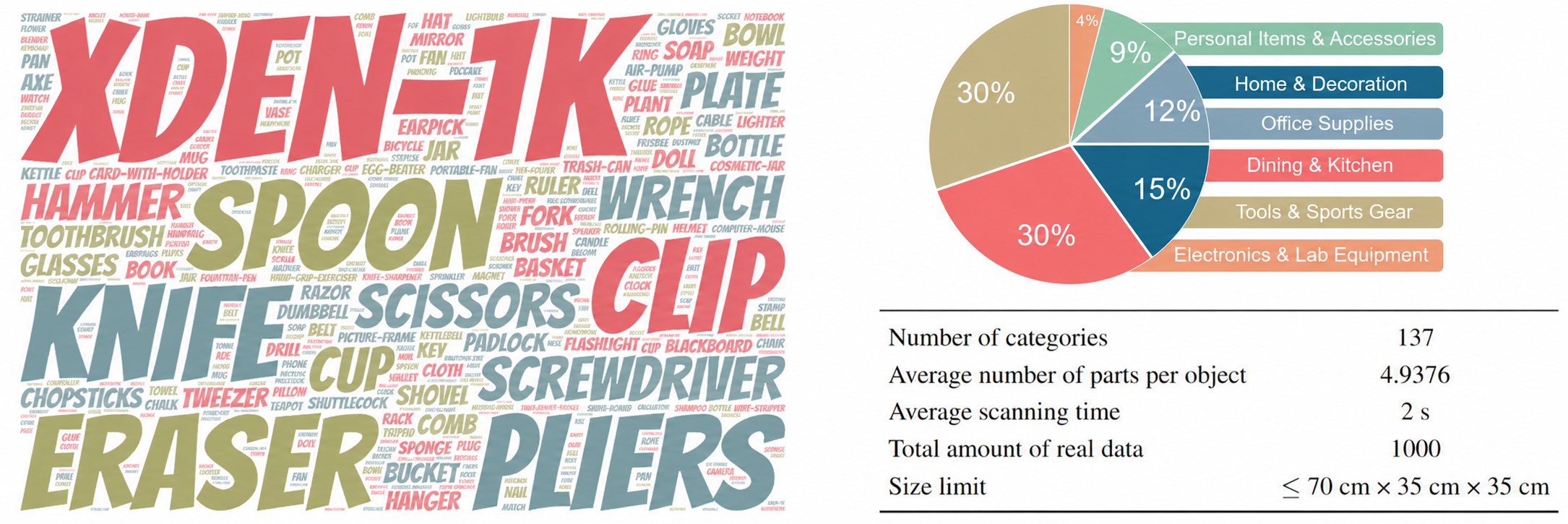}
    \caption{\textbf{Statistics of XDen-1K.}
Our dataset contains 1,000 everyday objects spanning 137 categories,
covering a diverse range of common items.}
    \label{fig:stats}
    \vspace{-1em}
\end{figure}


Substituting Eq.~\eqref{eq:region_mu} into Eq.~\eqref{eq:projection_integral}, the integral can be decomposed into region-wise path lengths:
\begin{equation}
\hat{I}_i(\mathbf{p}) =
I_0 \exp\!\left(- \sum_{k=1}^{K} \mu_k \, \ell_{i,k}(\mathbf{p}) \right),
\label{eq:forward_model_region}
\end{equation}
where $\ell_{i,k}(\mathbf{p}) = \int_{\mathcal{L}_i(\mathbf{p}) \cap \Omega_k} \mathrm{d}x$
denotes the path length of ray $\mathcal{L}_i(\mathbf{p})$ through region $\Omega_k$.  
This quantity can be precomputed geometrically from the known part segmentation and projection geometry.

The reconstruction task then reduces to estimating the material-wise attenuation vector
$\boldsymbol{\mu} = [\mu_1, \mu_2, \dots, \mu_K]^{\top}$.  
We jointly minimize the discrepancy between the simulated and measured projections from both biplanar views:
\begin{equation}
\hat{\boldsymbol{\mu}} =
\arg\min_{\boldsymbol{\mu}}
\sum_{i=1}^{2}
\sum_{\mathbf{p}}
\left|
I_i(\mathbf{p}; \boldsymbol{\mu})
- \tilde{I}_i(\mathbf{p})
\right|_2^2,
\label{eq:optimization}
\end{equation}
where $\tilde{I}_i(\mathbf{p})$ denotes the measured X-ray intensity on detector $i$.  
Each region (or part) is assigned one trainable parameter $\mu_k$, representing its LAC.  
The background region (air) is fixed as $\mu_{\text{air}}$.  
Using a differentiable forward-projection operator, the gradients
$\partial I_i(\mathbf{p}) / \partial \mu_k$
are computed automatically and the parameters $\boldsymbol{\mu}$ are optimized via Adam until convergence.

This formulation transforms the original high-dimensional inverse problem of estimating a dense field $\mu(\mathbf{x})$ into a low-dimensional optimization over a compact parameter set $\boldsymbol{\mu}$.  
The part-segmented mask serves as a powerful geometric prior that constrains the solution space, effectively regularizing the reconstruction and enabling tractable and physically consistent 3D recovery from only two biplanar X-ray projections.

\section{XDen-1K dataset}

\begin{figure}[t]
    \centering
    \includegraphics[width=1\linewidth]{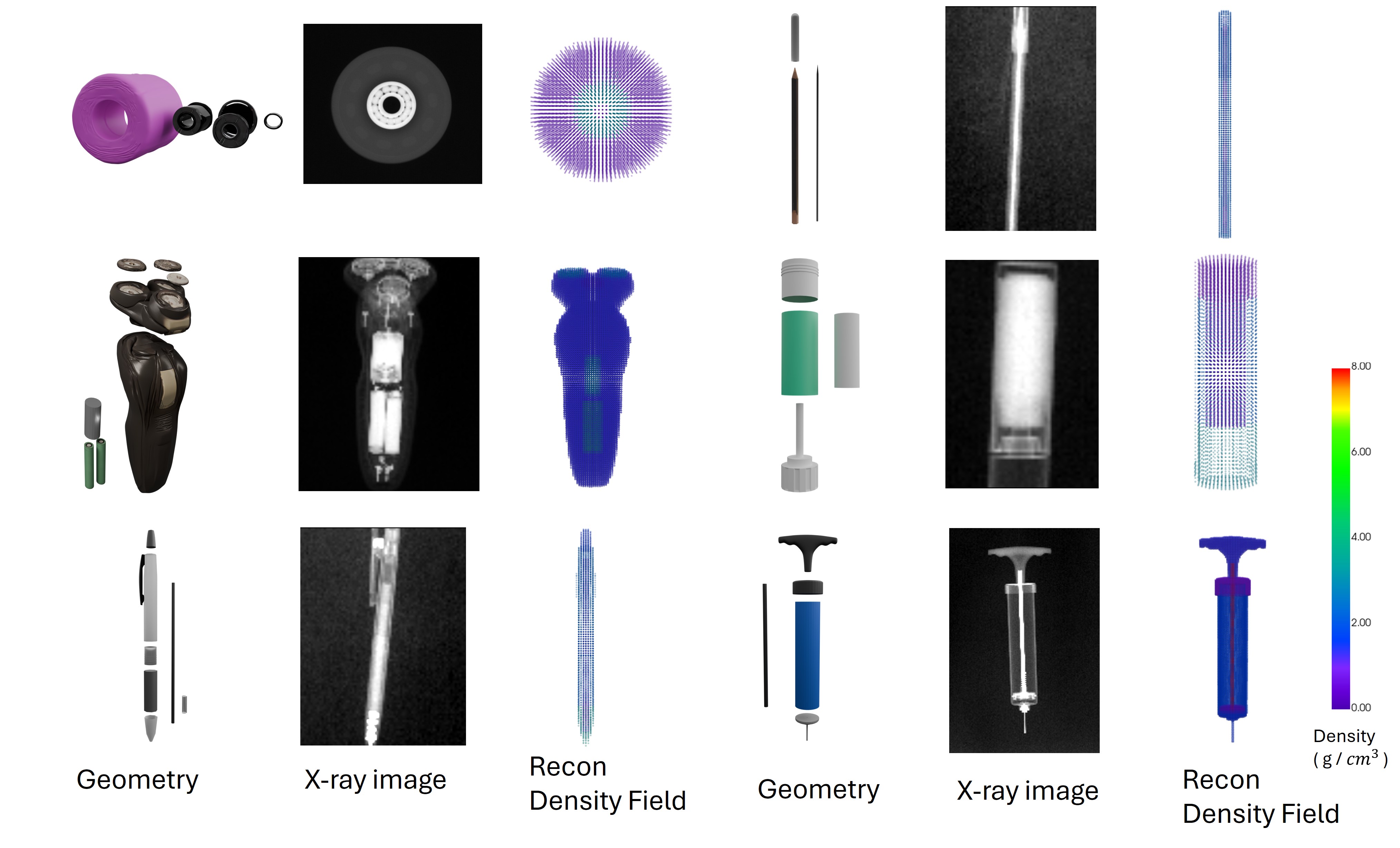}
    \vspace{-2em}
    \caption{\textbf{Gallery of XDen-1K.} XDen-1K dataset contains curated geometry, captured biplanar X-ray images, and reconstructed density fields. Some objects are represented by their original meshes, while others include manually modeled internal geometries.}
    \vspace{-2em}
    \label{fig:gallery}
\end{figure}
The reconstruction algorithm introduced in \cref{sec:recon} enables the recovery of a physically consistent volumetric density field. Building on this capability, a large-scale dataset, XDen-1K, was collected, comprising 1,000 real-world objects. It provides both real captured X-ray observations and corresponding reconstructed 3D density fields.

\subsection{Dataset Overview and Preprocessing}
To capture the diversity and complexity of objects encountered in everyday environments,
the objects covered in XDen-1K span \textbf{137} categories, ranging from tools and kitchenware
to consumer electronics. As shown in \cref{fig:stats}, these categories are relatively evenly distributed across common object families. Frequently appearing objects, such as spoons, screwdrivers, scissors and pliers, are all commonly encountered in daily life. This broad category coverage introduces rich variations in
\textbf{geometry, size, materials, and internal structure}, enabling the dataset to better
reflect real-world object variability.

Each object in the dataset is captured and annotated through a standardized pipeline, as shown in \cref{fig:pipeline}. We first collect calibrated multi-view RGB images to reconstruct a watertight mesh~\cite{zhang2024clay} and measure its real-world scale. A human-in-the-loop segmentation~\cite{Zhang2025BANGD3} is then performed to decompose objects into functionally meaningful components. To perform the density field reconstruction algorithm described in \cref{sec:recon}, ground-truth X-ray images are required as supervision. Therefore, for each object in XDen-1K, we also capture paired biplanar X-ray scans using our X-ray imaging machine. For some objects containing extremely complex interior geometry, we manually model out its interior structure based on the captured X-ray image which clearly reveals its inner structure.

\begin{table}[!htbp]
\centering
\vspace{-0.8em}
\caption{Mean ($\mu$) and standard deviation ($\sigma$) of reprojection errors on real X-rays.}
\label{tab:reproj_err}
\footnotesize
\renewcommand{\arraystretch}{0.9}
\begin{tabular*}{0.95\linewidth}{@{}l@{\extracolsep{\fill}}cccc@{}}
\toprule
Segmentation prior & LAT-$\mu$ & LAT-$\sigma$ & PA-$\mu$ & PA-$\sigma$ \\
\midrule
X-Field prediction & 1.121 & 0.287 & 0.724 & 0.268 \\
Current pipeline (manual labels) & 0.778 & 0.191 & 0.498 & 0.207 \\
Background noise & 0.220 & N/A & 0.200 & N/A \\
\bottomrule
\end{tabular*}
\vspace{-0.8em}
\end{table}

During actual biplanar X-ray capture stage, we use the Eagle Eye biplanar X-ray imaging system developed by TaoImage, with settings of  a 100 kV tube voltage, a 450 mA tube current, an exposure time of 280 ms per exposure, and a total exposure of 126 mAs.

Following this standardized acquisition pipeline, XDen-1K provides a comprehensive multi-modal representation for each object. Specifically, for each object, we provide a \textbf{part-segmented 3D mesh}, \textbf{biplanar X-ray scans}, and \textbf{a reconstructed volumetric density field}, as shown in \cref{fig:gallery}. For validation purposes, a subset of objects is additionally scanned using high-resolution CT, as described in \cref{sec:evalxden}.



\subsection{Evaluation of XDen-1K}
\label{sec:evalxden}

We evaluate the fidelity of the reconstructed density fields from two complementary perspectives: consistency with real X-ray observations and agreement with high-resolution CT references. First, to assess whether the optimized LAC field can faithfully explain the measured X-ray data, we reproject the reconstructed field using \cref{eq:projection_integral} and compare the resulting synthetic X-ray images with the real biplanar acquisitions. We report the Mean Absolute Error (MAE) and Pearson Correlation (Corr) in the image domain. As shown in \cref{tab:reproj_err}, our method achieves accurate and stable reprojection results when reliable geometric and structural priors are available, demonstrating the physical consistency of the reconstructed density fields.

For the CT-based evaluation, we note that the reference volumes are not perfect ground truth. Due to beam-hardening artifacts and the relatively low tube voltage, metallic components such as the air pump may suffer from streaking effects and deviate from the assumed LAC-to-density model in \cref{eq:estimate_rho}. We therefore use the CT scans only as partial volumetric references. As shown in \cref{tab:eval_comparison}, our reconstructions achieve reasonable agreement with these CT references, suggesting that the optimized density fields capture physically meaningful volumetric structure while remaining primarily constrained by the real X-ray observations.

\begin{table}[!htbp]
\centering
\caption{Quantitative evaluation of reconstruction quality on XDen-1K. Both density error against CT (MAE) and reprojection error against real X-ray images (LAT/PA) are measured. }
\setlength{\tabcolsep}{3.5pt} 
\vspace{-1em}
\begin{tabular}{l c cc cc cc cc}
\toprule
 & CT & \multicolumn{2}{c}{LAT} & \multicolumn{2}{c}{PA} & \multicolumn{2}{c}{Fine LAT} & \multicolumn{2}{c}{Fine PA} \\
 \cmidrule(lr){2-2} \cmidrule(lr){3-4} \cmidrule(lr){5-6} \cmidrule(lr){7-8} \cmidrule(lr){9-10}
 & MAE & MAE & Corr & MAE & Corr & MAE & Corr & MAE & Corr \\
\midrule
Glue    & 0.181 & 0.298 & 0.678 & 0.239 & 0.616 & 0.341 & 0.494 & 0.381 & 0.454 \\
Pencil  & 0.294 & 0.271 & 0.492 & 0.281 & 0.620 & 0.211 & 0.681 & 0.218 & 0.546 \\
AirPump & -     & 1.064 & 0.185 & 0.820 & 0.235 & 0.741 & 0.598 & 0.797 & 0.323 \\
BallPen & 0.710 & 0.286 & 0.065 & 0.258 & 0.410 & 0.239 & 0.382 & 0.330 & 0.228 \\
Razor   & 0.326 & 1.443 & 0.489 & 1.380 & 0.515 & 1.218 & 0.666 & 1.211 & 0.668 \\
\bottomrule
\end{tabular}
\label{tab:eval_comparison}
\vspace{-1em}
\end{table}

Manual interior-structure modeling can improve reconstruction quality but introduces additional effort that limits scalability. We therefore evaluate a more scalable variant that uses only external geometry and part segmentation as priors, while keeping the optimization procedure unchanged. As shown in \cref{tab:eval_comparison}, removing manually curated interior priors causes an expected accuracy drop, but the method still maintains reasonable reprojection consistency and volumetric fidelity, demonstrating robustness under weaker priors.



\section{Applications on XDen-1K}
Building on XDen-1K, we demonstrate three downstream applications: benchmarking existing physics-reasoning methods, \textbf{X-ray-conditioned volumetric segmentation}, and \textbf{center-of-mass-aware robot manipulation}. We argue that X-ray imaging is a key modality for spatial perception, as it provides rich cues about object geometry, internal structure, and material distribution. When paired with volumetric density field, XDen-1K enables physically grounded object reasoning and supports a broad range of downstream tasks.

\subsection{Benchmarking Physical Reasoning Methods}

Reasoning about physical properties is fundamental to building physically grounded digital worlds and enabling stable robot manipulation. Among these properties, density is particularly important, as it can be directly converted into mass-related quantities such as total mass, center of mass, and moments of inertia. Given accurate geometry and curated object structures, XDen-1K provides \textbf{high-fidelity, physically calibrated} volumetric density field reconstructed by our pipeline, with quantitative validation presented in \cref{sec:evalxden}.

This makes XDen-1K a suitable benchmark for supervising and evaluating density estimation methods. We evaluate recent methods, including Pixie~\cite{Le2025PixieFA}, GaussianProperty~\cite{Xu2024GaussianPropertyIP}, and NeRF2Physics~\cite{Zhai2024PhysicalPU}, using voxel-wise density error, L1 center-of-mass distance, and mean absolute percentage error (MAPE) of total mass. Results are summarized in \cref{tab:phys_comp}. LLM-based methods, NeRF2Physics achieves the best performance in density MAE and center-of-mass distance, while learning-based methods, Pixie obtains the lowest mass MAPE.

\subsection{X-ray-conditioned volumetric segmentation}
X-ray images provide strong cues about material composition and internal structure, making them effective conditioning signals for volumetric segmentation. We therefore extend PartField~\cite{Liu2025PARTFIELDL3} to the volumetric domain and introduce \textbf{X-Field}, which conditions on biplanar X-ray images to segment an object's 3D volume. X-Field is trained on a synthetic dataset built upon PartNeXt~\cite{Wang2025PartNeXtAN}, with material properties synthesized to match our real-world collection. Experiments on both synthetic and real data show that X-Field consistently outperforms PartField, validating the benefit of X-ray conditioning for volumetric inference.

\subsubsection{Conditioning X-ray modality on PartField}
PartField~\cite{Liu2025PARTFIELDL3} takes an input point cloud $(P)$, extracts per-point features with PVCNN~\cite{liu2019pvcnn}, and lifts them into a triplane representation. A transformer decoder predicts triplane features $(L_{\text{tri}})$, which define a queryable feature field $(F_{\text{tri}}(\cdot))$ in 3D space and can be projected back onto the input points to obtain point-wise features.

We augment this backbone with X-ray conditioning. Given biplanar X-ray images $(I_X)$, an image encoder extracts X-ray features $(L_X = E_X(I_X))$. Since X-ray appearance is strongly scale-dependent, we additionally provide the physical scale $(\lambda)$ and use FiLM~\cite{perez2017filmvisualreasoninggeneral} to modulate the encoded X-ray features into a scale-aware latent representation:
\begin{equation}
L_X^{\lambda} = \operatorname{FiLM}(L_X, \lambda).
\label{eq:xray_film}
\end{equation}
The resulting scale-aware X-ray features are fused with the triplane features through cross-attention:
\begin{equation}
\tilde{L}_{\text{tri}}
= \operatorname{CrossAttention}\bigl(L_{\text{tri}}, L_X^{\lambda}\bigr).
\label{eq:fuse}
\end{equation}
This produces an X-ray-conditioned volumetric feature field $(\tilde{F}_{\text{tri}}(\cdot))$.

\begin{table}[t]
\centering
\vspace{-1.1em}
\caption{Comparison of physical reasoning methods in terms of density, center of mass (CoM), and total mass.}
\label{tab:phys_comp}
\footnotesize
\renewcommand{\arraystretch}{0.9}
\begin{tabular*}{0.95\linewidth}{@{\extracolsep{\fill}}lccc}
\toprule
Method & Density MAE & CoM distance (cm) & Mass MAPE \\
\midrule
NeRF2Physics      & 1.157 & 0.523 & 1.174 \\
GaussianProperty  & 1.295 & 0.632 & 1.247 \\
Pixie             & 1.228 & 0.547 & 0.713 \\
\bottomrule
\end{tabular*}
\vspace{-1.0em}
\end{table}

Our goal is to predict a dense volumetric density field rather than only point-wise features on the input surface. We therefore randomly sample query points $(P_{\text{query}})$ within the object volume and query the fused triplane representation at these locations. For each query point $(p \in P_{\text{query}})$, a density MLP head regresses its density value:
\begin{equation}
\hat{D}(p) = f_{\theta}\bigl(\tilde{F}_{\text{tri}}(p)\bigr).
\label{eq:density_pred}
\end{equation}
Training is supervised with an $(\ell_2)$ loss between the predicted densities and the ground-truth densities from the synthetic volumetric fields:
\begin{equation}
\mathcal{L}_{\text{den}}
=
\frac{1}{|P_{\text{query}}|}
\sum_{p \in P_{\text{query}}}
\left\|
\hat{D}(p) - D(p)
\right\|_2^2 .
\label{eq:density_loss}
\end{equation}
\begin{figure}[t]
    \centering
    \includegraphics[width=\linewidth]{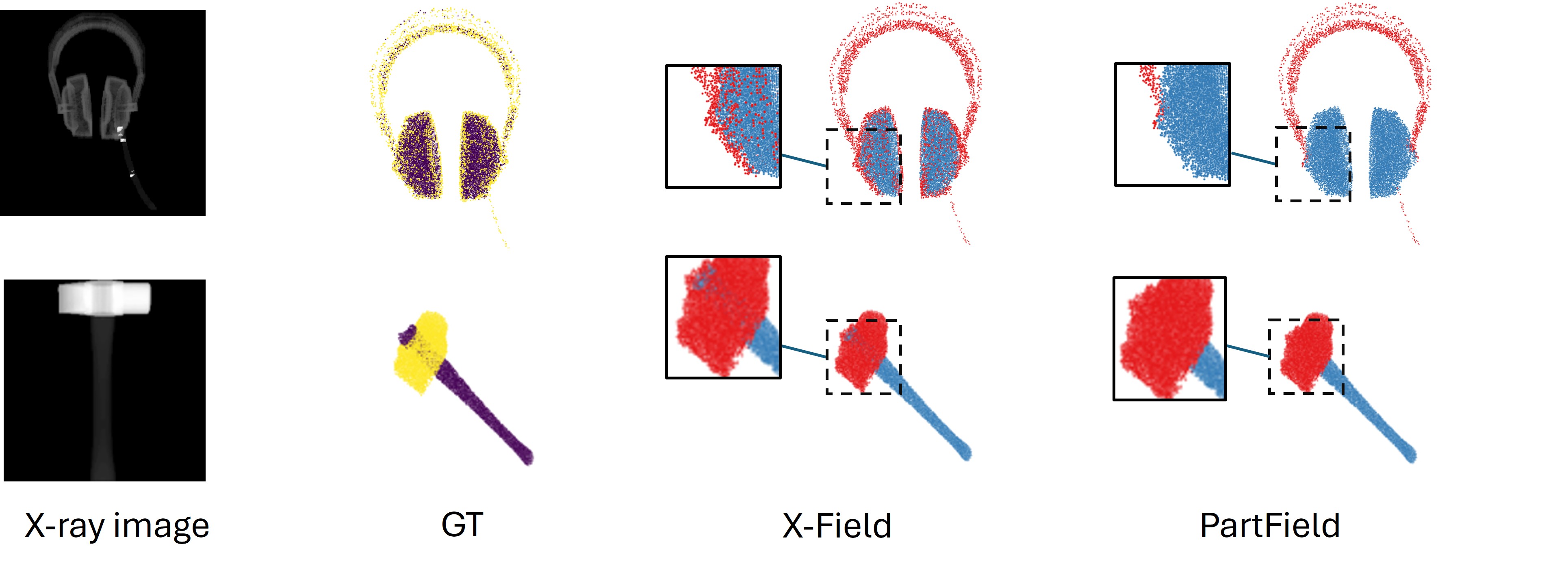}
    \caption{Comparison of qualitative segmentation results. }
    \label{fig:xfield_result}
\end{figure}

\begin{table}[t]
    \centering
    \footnotesize
    \renewcommand{\arraystretch}{1.15}
    \begin{tabular*}{0.95\linewidth}{@{\extracolsep{\fill}}lcc}
        \toprule
        \textbf{Method} & \textbf{Synthetic Data IoU $\uparrow$} & \textbf{Real Data IoU $\uparrow$} \\
        \midrule
        PartField      & 53.185 & 38.410 \\
        X-Field (Ours) & \textbf{72.645} & \textbf{49.251} \\
        \bottomrule
    \end{tabular*}

    \vspace{1em}
    \caption{Comparison of IoU scores on synthetic and real data.}
    \label{tab:xfield_iou}
    \vspace{-2em}
\end{table}

Biplanar X-ray inputs may suffer from missing-view ambiguity due to limited viewing directions. To mitigate this issue, we introduce an auxiliary reconstruction objective: an additional X-ray image is decoded from features associated with the third plane and supervised using an MSE loss against the corresponding ground-truth X-ray image. This auxiliary task regularizes 3D feature learning and encourages more complete volumetric reasoning. Finally, by applying $(k)$-means clustering to the predicted densities at the query points, the volume is partitioned into regions with similar density values, producing a material-aware volumetric segmentation.

\subsubsection{Evaluation on  X-ray-conditioned segmentation}

The volumetric segmentation performance of the original PartField and the proposed X-Field is evaluated on both synthetic and real data. As shown in \cref{fig:xfield_result} and \cref{tab:xfield_iou}, conditioning on biplanar X-ray images leads to both higher IoU scores and better volumetric segmentations.

\noindent\textbf{Toward Fully Automatic Dataset Construction.}
Although X-Field shows encouraging real-world performance, \cref{tab:reproj_err} shows that its predicted priors still produce higher reprojection errors than manual labels. We therefore use manually curated labels as ground truth in the current XDen-1K pipeline and present X-Field as a promising downstream application. With more high-quality synthetic labels and X-ray-consistent material properties, X-Field could further improve sim-to-real transfer and accelerate future dataset acquisition.

\subsection{CoM-aware Robot Manipulation}

\begin{figure*}[t]
    \centering
    \includegraphics[page=1,width=0.95\textwidth]{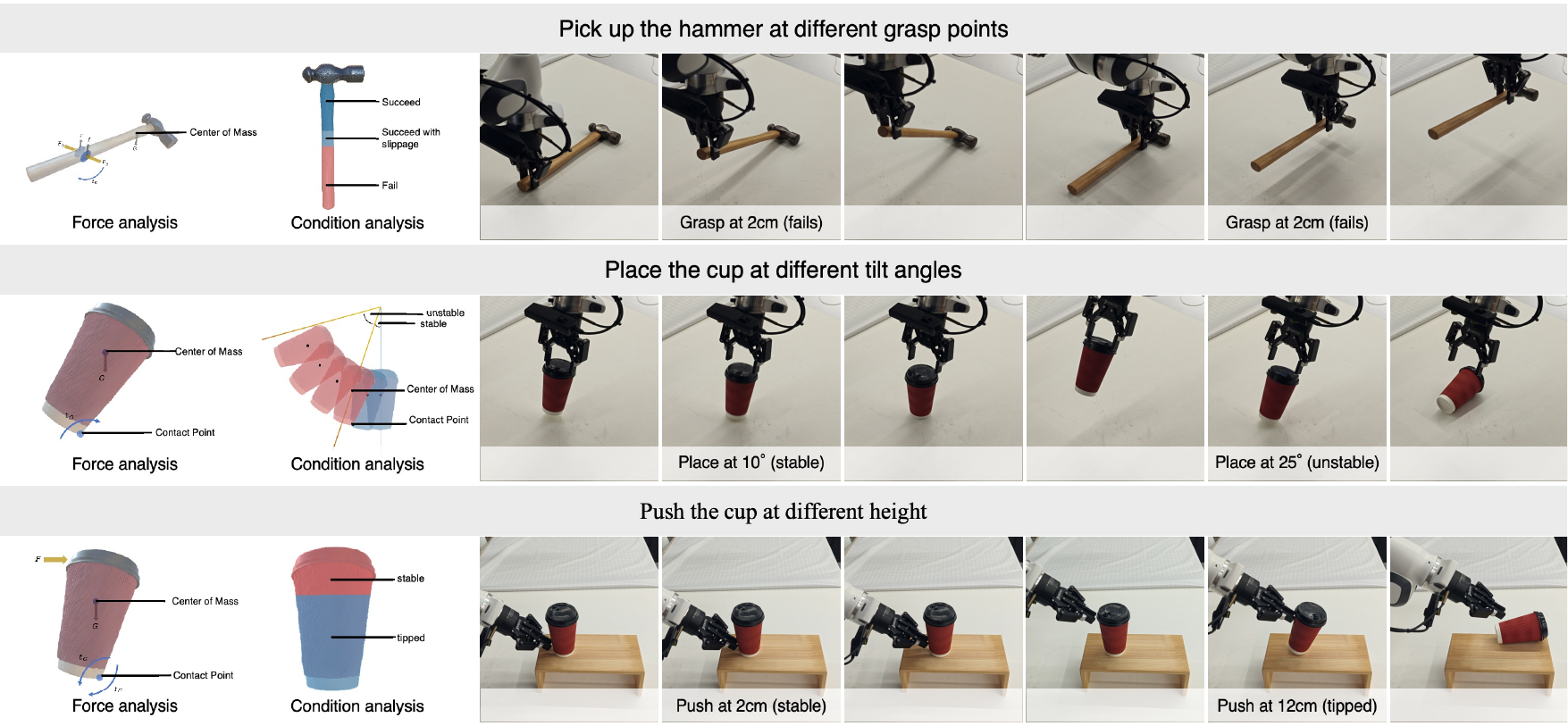}
    \caption{Effect of the object's center of mass on manipulation stability under Pick, Place and Push task. 
            Left: force and stability analysis. Right: real-robot executions.}
    \label{fig:real_exp}
\vspace{-1em}
\end{figure*}

The perception of an object's center of mass is fundamental for achieving physically stable and dynamically consistent manipulation. 
Beyond geometric reasoning, understanding the relationship between the center of mass, external forces, and contact points allows a robot to anticipate rotation, slippage, and tipping during interaction.
To study how the center of mass influences manipulation outcomes, we conduct three representative experiments on a Franka Emika Panda robotic arm equipped with a parallel gripper, as shown in \cref{fig:real_exp}.

\textbf{Pick.} Grasping near the CoM produces a small gravitational moment, enabling stable lifting. Increasing the grasp offset enlarges the lever arm and torque, which can overwhelm contact friction, induce rotation, and cause grasp failure.

\textbf{Place.} Stability is governed by whether the CoM projection remains inside the support region. Small tilts create a restoring moment and the object self-rights; larger tilts move the projection beyond the support boundary and result in tipping.

\textbf{Push.} The pushing force creates a moment about the CoM. Forces applied below the CoM favor translation, whereas forces applied above the CoM generate a larger overturning moment and can tip the object.

Beyond qualitative demonstrations, the benefit of CoM reasoning is further evaluated in a downstream grasping pipeline. Density-field prior is used to guide grasp selection in AnyGrasp~\cite{fang2023anygrasp} by selecting the candidate grasp pose whose gripper endpoints are closest to the predicted CoM. Grasp stability is evaluated using grasp success rate and joint torque loads, with particular attention to wrist torque (Joint~7), which reflects the end-effector load during grasp execution. Results are compared across three settings: vanilla AnyGrasp, AnyGrasp guided by the NeRF2Physics CoM, and AnyGrasp guided by the CoM from our reconstructed result. As shown in \cref{tab:robotic_success_rate}, accurate CoM guidance from the proposed method substantially improves grasp success rate while reducing both wrist and mean joint torques, indicating more stable and mechanically efficient grasps

These experiments indicate that reasoning about the object's center of mass plays an important role in improving related robotic manipulation task. Leveraging this physical understanding from our dataset can contribute to more stable behaviors and more reliable interaction with diverse objects in real-world settings.

\vspace{-1em}
\begin{table}[htbp]
\centering
\caption{Grasp success (\%) and joint torque ($\mathrm{N\cdot m}$) under CoM guidance.}
\label{tab:robotic_success_rate}
\renewcommand{\arraystretch}{1}
\setlength{\tabcolsep}{8pt}
\begin{tabular}{lccc}
\toprule
& Success Rate  $\uparrow$ & Joint-7 torque $\downarrow$  & Mean torque $\downarrow$ \\
\midrule
AnyGrasp & 27.50 & 0.521 & 9.645 \\
+NeRF2Physics CoM & 55.00 &0.443 & 9.529 \\
+Ours CoM & \textbf{77.75}&\textbf{0.435}&\textbf{9.497}\\
\bottomrule
\end{tabular}
\end{table}
\vspace{-1em}
%


\section{Conclusion}

We presented  \textbf{XDen-1K}, a large-scale real-world dataset comprising paired biplanar X-ray scans and reconstructed 3D density fields. A differentiable X-ray optimization framework is developed to leverage shape and segmentation priors for recovering piecewise-constant densities from two X-ray views, thereby transforming an otherwise ill-posed problem into a tractable reconstruction task. Evaluations against CT references and measured X-ray images show that the proposed method can recover high-fidelity density fields across a wide range of objects.
Beyond the dataset and reconstruction framework, XDen-1K introduces three subtasks: benchmarking density estimation methods, X-ray-conditioned volumetric segmentation, and center-of-mass-aware robotic manipulation. The results establish XDen-1K as a physically grounded benchmark and reference resource for physical property estimation, while highlighting the potential of X-ray imaging as a complementary modality for 3D segmentation and object understanding. With more robust segmentation models and standardized X-ray acquisition protocols, the data acquisition pipeline can be further automated and scaled, enabling broader applications in physical reasoning and embodied AI.

\section{Acknowledgement}
This work was supported in part by the National Natural Science Foundation of China under Grant (W2431046), Central Guided Local Science and Technology Foundation of China (YDZX20253100001001), Shanghai Pujiang Program (24PJA080) and by MoE Key Lab of Intelligent Perceptionand Human-Machine Collaboration (ShanghaiTech University), the Shanghai Frontiers Science Center of Human-centered Artificial Intelligence. The experiments of this work were supported by the SIST computing Platform and HPC, ShanghaiTech University.
We gratefully acknowledge the invaluable discussion and feedback provided by Peijun Xu, Qixuan Zhang and Longwen Zhang from ShanghaiTech University.
%
%
\bibliographystyle{splncs04}
\bibliography{main}

@String{Computer = "{IEEE} Computer" }

@STRING{CVPR = {CVPR}}

@STRING{ECCV = {ECCV}}

@STRING{ICCV = {ICCV}}

@STRING{ACCV = {ACCV}}

@STRING{NEURIPS = {NEURIPS}}

@STRING{ICLR = {ICLR}}

@STRING{ICASSP = {ICASSP}}

@STRING{SIGGRAPH = {SIGGRAPH Conference Papers}}

@STRING{TOG = {ACM Trans. on Graphics}}

@STRING{ICRA = {ICRA}}

@STRING{AAAI = {AAAI}}

@STRING{COMPUTER = {Computer}}

@STRING{ACM = {ACM}}

@STRING{ARXIV = {arXiv.org}}

@STRING{NETWORKS = {Networks}}

@STRING{CHI = {CHI}}

@STRING{RIC = {Readings in Computer Vision: Issues, Problems,
Principles, and Paradigms (RIC)}}

@STRING{Nature = {Nature}}

@inproceedings{NEURIPS2024_d7af02c8,
 author = {Guo, Minghao and Wang, Bohan and Ma, Pingchuan and Zhang, Tianyuan and Owens, Crystal Elaine and Gan, Chuang and Tenenbaum, Joshua B. and He, Kaiming and Matusik, Wojciech},
 booktitle = NEURIPS,
 title = {Physically Compatible 3D Object Modeling from a Single Image},
 year = {2024}
}

@article{fang2023anygrasp,
  title={Anygrasp: Robust and efficient grasp perception in spatial and temporal domains},
  author={Fang, Hao-Shu and Wang, Chenxi and Fang, Hongjie and Gou, Minghao and Liu, Jirong and Yan, Hengxu and Liu, Wenhai and Xie, Yichen and Lu, Cewu},
  journal={IEEE Transactions on Robotics},
  year={2023},
  publisher={IEEE}
}

@inproceedings{tung2023physionevaluatingphysicalscene,
      title={Physion++: Evaluating Physical Scene Understanding that Requires Online Inference of Different Physical Properties}, 
      author={Hsiao-Yu Tung and Mingyu Ding and Zhenfang Chen and Daniel Bear and Chuang Gan and Joshua B. Tenenbaum and Daniel LK Yamins and Judith E Fan and Kevin A. Smith},
      year={2023},
      booktitle=NEURIPS
}

@inproceedings{liu2019pvcnn,
  title={Point-Voxel CNN for Efficient 3D Deep Learning},
  author={Liu, Zhijian and Tang, Haotian and Lin, Yujun and Han, Song},
  booktitle=NEURIPS,
  year={2019}
}

@inproceedings{shuai2025pugs,
  title={Pugs: Zero-shot physical understanding with gaussian splatting},
  author={Shuai, Yinghao and Yu, Ran and Chen, Yuantao and Jiang, Zijian and Song, Xiaowei and Wang, Nan and Zheng, Jv and Ma, Jianzhu and Yang, Meng and Wang, Zhicheng and others},
  booktitle=ICRA,
  year={2025}
}

@inproceedings{huang2024materialanythinggeneratingmaterials,
      title={Material Anything: Generating Materials for Any 3D Object via Diffusion}, 
      author={Xin Huang and Tengfei Wang and Ziwei Liu and Qing Wang},
      year={2024},
      booktitle=CVPR
}

@article{mildenhall2020nerfrepresentingscenesneural,
  title={Nerf: Representing scenes as neural radiance fields for view synthesis},
  author={Mildenhall, Ben and Srinivasan, Pratul P and Tancik, Matthew and Barron, Jonathan T and Ramamoorthi, Ravi and Ng, Ren},
  journal={Communications of the ACM},
  year={2021}
}

@inproceedings{jiang2025phystwinphysicsinformedreconstructionsimulation,
  title={Phystwin: Physics-informed reconstruction and simulation of deformable objects from videos},
  author={Jiang, Hanxiao and Hsu, Hao-Yu and Zhang, Kaifeng and Yu, Hsin-Ni and Wang, Shenlong and Li, Yunzhu},
  booktitle=ICCV,
  year={2025}
}

@article{10.1145/3618358,
author = {Ma, Xiaohe and Xu, Xianmin and Zhang, Leyao and Zhou, Kun and Wu, Hongzhi},
title = {OpenSVBRDF: A Database of Measured Spatially-Varying Reflectance},
year = {2023},
journal = TOG
}

@inproceedings{ahmed20253dcompat200languagegroundedcompositionalunderstanding,
  title={3DCoMPaT200: Language-grounded compositional understanding of parts and materials of 3D shapes},
  author={Ahmed, Mahmoud and Li, Xiang and Prajapati, Arpit and Elhoseiny, Mohamed},
  journal=NEURIPS,
  year={2024}
}

@article{Zhang2025BANGD3,
    title={BANG: Dividing 3D Assets via Generative Exploded Dynamics},
    year={2025},
    journal=TOG,
    author={Zhang, Longwen and Zhang, Qixuan and Jiang, Haoran and
    Bai, Yinuo and Yang, Wei and Xu, Lan and Yu, Jingyi}
}

@article{glover2016objectively,
  title={An objectively-analyzed method for measuring the useful penetration of x-ray imaging systems},
  author={Glover, Jack L and Hudson, Lawrence T},
  journal={Measurement Science and Technology},
  year={2016}
}

@article{feldkamp1984practical,
  title={Practical cone-beam algorithm},
  author={Feldkamp, Lee A and Davis, Lloyd C and Kress, James W},
  journal={Journal of the Optical Society of America A},
  year={1984}
}

@article{slota2012stability,
  title={Stability control of grasping objects with different locations of center of mass and rotational inertia},
  author={Slota, Gregory P and Suh, Moon Suk and Latash, Mark L and Zatsiorsky, Vladimir M},
  journal={Journal of motor behavior},
  year={2012},
}

@article{rodet2004cone,
  title={The cone-beam algorithm of Feldkamp, Davis, and Kress preserves oblique line integrals},
  author={Rodet, Thomas and Noo, Fr{\'e}d{\'e}ric and Defrise, Michel},
  journal={Medical physics},
  year={2004}
}

@article{lin2014efficient,
  title={An efficient polyenergetic SART (pSART) reconstruction algorithm for quantitative myocardial CT perfusion},
  author={Lin, Yuan and Samei, Ehsan},
  journal={Medical physics},
  year={2014},
}

@article{nuyts2013modelling,
  title={Modelling the physics in the iterative reconstruction for transmission computed tomography},
  author={Nuyts, Johan and De Man, Bruno and Fessler, Jeffrey A and Zbijewski, Wojciech and Beekman, Freek J},
  journal={Physics in Medicine \& Biology},
  year={2013}
}

@article{sun2022review,
  title={Review of high energy x-ray computed tomography for non-destructive dimensional metrology of large metallic advanced manufactured components},
  author={Sun, Wenjuan and Symes, Daniel R and Brenner, Ceri M and B{\"o}hnel, Michael and Brown, Stephen and Mavrogordato, Mark N and Sinclair, Ian and Salamon, Michael},
  journal={Reports on Progress in Physics},
  year={2022}
}

@article{ginat2014advances,
  title={Advances in computed tomography imaging technology},
  author={Ginat, Daniel Thomas and Gupta, Rajiv},
  journal={Annual review of biomedical engineering},
  year={2014},
}

@inproceedings{cai2024structure,
  title={Structure-aware sparse-view x-ray 3d reconstruction},
  author={Cai, Yuanhao and Wang, Jiahao and Yuille, Alan and Zhou, Zongwei and Wang, Angtian},
  booktitle=CVPR,
  year={2024}
}

@article{wu2023self,
  title={Self-supervised coordinate projection network for sparse-view computed tomography},
  author={Wu, Qing and Feng, Ruimin and Wei, Hongjiang and Yu, Jingyi and Zhang, Yuyao},
  journal={IEEE Transactions on Computational Imaging},
  year={2023},
}

@article{berg2020experiences,
  title={Experiences with a new biplanar low-dose X-ray device for imaging the facial skeleton: A feasibility study},
  author={Berg, Britt-Isabelle and Laville, Aur{\'e}lien and Courvoisier, Delphine S and Rouch, Philippe and Schouman, Thomas},
  journal={PloS one},
  year={2020},
  publisher={Public Library of Science San Francisco, CA USA}
}

@article{chen2024automatic,
  title={Automatic 3D reconstruction of vertebrae from orthogonal bi-planar radiographs},
  author={Chen, Yuepeng and Gao, Yue and Fu, Xiangling and Chen, Yingyin and Wu, Ji and Guo, Chenyi and Li, Xiaodong},
  journal={Scientific Reports},
  year={2024},
  publisher={Nature Publishing Group UK London}
}

@inproceedings{ying2019x2ct,
  title={X2CT-GAN: reconstructing CT from biplanar X-rays with generative adversarial networks},
  author={Ying, Xingde and Guo, Heng and Ma, Kai and Wu, Jian and Weng, Zhengxin and Zheng, Yefeng},
  booktitle=CVPR,
  year={2019}
}

@inproceedings{jeong2025dx2ct,
  title={DX2CT: Diffusion Model for 3D CT Reconstruction from Bi or Mono-planar 2D X-ray (s)},
  author={Jeong, Yun Su and Yoo, Hye Bin and Chun, Il Yong},
  booktitle=ICASSP,
  year={2025}
}

@inproceedings{kyung2023perspective,
  title={Perspective projection-based 3d ct reconstruction from biplanar x-rays},
  author={Kyung, Daeun and Jo, Kyungmin and Choo, Jaegul and Lee, Joonseok and Choi, Edward},
  booktitle=ICASSP,
  year={2023}
}

@inproceedings{huang2024generalizable,
  title={Generalizable Structure-Aware INF: Biplanar-View CT Reconstruction via Disentangled Implicit Neural Field},
  author={Huang, Bei and Pei, Yuru},
  booktitle=ACCV,
  year={2024}
}

@article{melhem2016eos,
  title={EOS{\textregistered} biplanar X-ray imaging: concept, developments, benefits, and limitations},
  author={Melhem, Elias and Assi, Ayman and El Rachkidi, Rami and Ghanem, Ismat},
  journal={Journal of children's orthopaedics},
  year={2016}
}

@techreport{berger1987xcom,
  title={XCOM: Photon cross sections on a personal computer},
  author={Berger, Martin J and Hubbell, John Howard},
  year={1987},
  institution={National Bureau of Standards, Washington, DC (USA). Center for Radiation~…}
}

@inproceedings{zhou2025pointsam,
    title={Point-{SAM}: Promptable 3D Segmentation Model for Point Clouds},
    author={Yuchen Zhou and Jiayuan Gu and Tung Yen Chiang and Fanbo
    Xiang and Hao Su},
    booktitle= ICLR,
    year={2025},
}

@inproceedings{Liu2025PARTFIELDL3,
    title={PartField: Learning 3D Feature Fields for Part
    Segmentation and Beyond},
    author={Minghua Liu and Mikaela Angelina Uy and Donglai Xiang and
    Hao Su and Sanja Fidler and Nicholas Sharp and Jun Gao},
    year={2025},
    booktitle=ICCV
}

@inproceedings{Wang2025PartNeXtAN,
  title={PartNeXt: A Next-Generation Dataset for Fine-Grained and Hierarchical 3D Part Understanding},
  author={Penghao Wang and Yiyang He and Xin Lv and Yukai Zhou and Lan Xu and Jingyi Yu and Jiayuan Gu},
  year={2025},
  booktitle=NEURIPS
}

@inproceedings{Cao2025PhysX3DP3,
  title={PhysX-3D: Physical-Grounded 3D Asset Generation},
  author={Ziang Cao and Zhaoxi Chen and Liang Pan and Ziwei Liu},
  booktitle=NEURIPS,
  year={2025}
}

@article{Le2025PixieFA,
  title={Pixie: Fast and Generalizable Supervised Learning of 3D Physics from Pixels},
  author={Long Le and Ryan Lucas and Chen Wang and Chuhao Chen and Dinesh Jayaraman and Eric Eaton and Lingjie Liu},
  journal={ArXiv},
  year={2025},
  volume={abs/2508.17437}
}

@inproceedings{Xu2024GaussianPropertyIP,
  title={Gaussianproperty: Integrating physical properties to 3d gaussians with lmms},
  author={Xu, Xinli and Ge, Wenhang and Qiu, Dicong and Chen, ZhiFei and Yan, Dongyu and Liu, Zhuoyun and Zhao, Haoyu and Zhao, Hanfeng and Zhang, Shunsi and Liang, Junwei and others},
  booktitle=ICCV,
  year={2025}
}

@inproceedings{Zhai2024PhysicalPU,
  title={Physical Property Understanding from Language-Embedded Feature Fields},
  author={Albert J. Zhai and Yuan Shen and Emily Y. Chen and Gloria X. Wang and Xinlei Wang and Sheng Wang and Kaiyu Guan and Shenlong Wang},
  booktitle=CVPR,
  year={2024}
}

@inproceedings{huang2024dreamphysics,
  title={DreamPhysics: Learning Physical Properties of Dynamic 3D Gaussians with Video Diffusion Priors},
  author={Huang, Tianyu and Zeng, Yihan and Li, Hui and Zuo, Wangmeng and Lau, Rynson WH},
  booktitle=AAAI,
  year={2024}
}

@article{Xie2025Vid2SimRA,
  title={Vid2Sim: Realistic and Interactive Simulation from Video for Urban Navigation},
  author={Ziyang Xie and Zhizheng Liu and Zhenghao Peng and Wayne Wu and Bolei Zhou},
  booktitle=CVPR,
  year={2025},
  pages={1581-1591}
}

@article{Li2023PACNeRFPA,
  title={PAC-NeRF: Physics Augmented Continuum Neural Radiance Fields for Geometry-Agnostic System Identification},
  author={Xuan Li and Yi-Ling Qiao and Peter Yichen Chen and Krishna Murthy Jatavallabhula and Ming Lin and Chenfanfu Jiang and Chuang Gan},
  journal={arXiv preprint arXiv:2303.05512},
  year={2023}
}

@inproceedings{cao2025physxanythingsimulationreadyphysical3d,
      title={PhysX-Anything: Simulation-Ready Physical 3D Assets from Single Image}, 
      author={Ziang Cao and Fangzhou Hong and Zhaoxi Chen and Liang Pan and Ziwei Liu},
      year={2026},
      booktitle=CVPR
}

@inproceedings{perez2017filmvisualreasoninggeneral,
  title={Film: Visual reasoning with a general conditioning layer},
  author={Perez, Ethan and Strub, Florian and De Vries, Harm and Dumoulin, Vincent and Courville, Aaron},
  booktitle=AAAI,
  year={2018}
}

@inproceedings{CLEVRER2020ICLR,
  author    = {Kexin Yi and
               Chuang Gan and
               Yunzhu Li and
               Pushmeet Kohli and
               Jiajun Wu and
               Antonio Torralba and
               Joshua B. Tenenbaum},
  title     = {{CLEVRER:} Collision Events for Video Representation and Reasoning},
  booktitle = {ICLR},
  year      = {2020}
}

@article{yao2025castcomponentaligned3dscene,

      title={CAST: Component-Aligned 3D Scene Reconstruction from an RGB Image}, 

      author={Kaixin Yao and Longwen Zhang and Xinhao Yan and Yan Zeng and Qixuan Zhang and Lan Xu and Wei Yang and Jiayuan Gu and Jingyi Yu},

      year={2025},

      journal=TOG

}

@inproceedings{bakhtin2019phyre,
  title     = {PHYRE: A New Benchmark for Physical Reasoning},
  author    = {Bakhtin, Anton and van der Maaten, Laurens and Johnson, Justin and Gustafson, Laura and Girshick, Ross},
  booktitle = NEURIPS,
  year      = {2019}
}

@inproceedings{Li2024IPHYRE,
  author    = {Shiqian Li and Kewen Wu and Chi Zhang and Yixin Zhu},
  title     = {I-PHYRE: Interactive Physical Reasoning},
  booktitle = ICLR,
  year      = {2024},
}

@inproceedings{xiang2024structured,
    title   = {Structured 3D Latents for Scalable and Versatile 3D Generation},
    author  = {Xiang, Jianfeng and Lv, Zelong and Xu, Sicheng and Deng, Yu and Wang, Ruicheng and Zhang, Bowen and Chen, Dong and Tong, Xin and Yang, Jiaolong},
    booktitle=CVPR,
    year    = {2024}
}

@inproceedings{Chang2025Winding,
  title     = {Lifting the Winding Number: Precise Discontinuities in Neural Fields for Physics Simulation},
  author    = {Yue Chang and Mengfei Liu and Zhecheng Wang and Peter Yichen Chen and Eitan Grinspun},
  booktitle = SIGGRAPH,
  year      = {2025}
}

@article{kumar2019scalablemodularmaterialpoint,
  title={Scalable and modular material point method for large-scale simulations},
  author={Kumar, Krishna and Salmond, Jeffrey and Kularathna, Shyamini and Wilkes, Christopher and Tjung, Ezra and Biscontin, Giovanna and Soga, Kenichi},
  journal={arXiv preprint arXiv:1909.13380},
  year={2019}
}

@inproceedings{wang2025vggtvisualgeometrygrounded,
      title={VGGT: Visual Geometry Grounded Transformer}, 
      author={Jianyuan Wang and Minghao Chen and Nikita Karaev and Andrea Vedaldi and Christian Rupprecht and David Novotny},
      year={2025},
      booktitle=CVPR, 
}

@inproceedings{zhong2024meshsegmenter,
    title={Meshsegmenter: Zero-shot mesh semantic segmentation via
    texture synthesis},
    author={Zhong, Ziming and Xu, Yanyu and Li, Jing and Xu, Jiale
    and Li, Zhengxin and Yu, Chaohui and Gao, Shenghua},
    booktitle=ECCV,
    year={2024}
}

@inproceedings{ma20243d,
    title={Find any part in 3d},
    author={Ma, Ziqi and Yue, Yisong and Gkioxari, Georgia},
    booktitle=ICCV,
    year={2025}
}

@article{ma2025p3sam,
    title={P3-sam: Native 3d part segmentation},
    author={Ma, Changfeng and Li, Yang and Yan, Xinhao and Xu,
        Jiachen and Yang, Yunhan and Wang, Chunshi and Zhao, Zibo and
    Guo, Yanwen and Chen, Zhuo and Guo, Chunchao},
    journal={arXiv preprint arXiv:2509.06784},
    year={2025}
}

@article{yang2024sampart3d,
    title={Sampart3d: Segment any part in 3d objects},
    author={Yang, Yunhan and Huang, Yukun and Guo, Yuan-Chen and Lu,
        Liangjun and Wu, Xiaoyang and Lam, Edmund Y and Cao, Yan-Pei and
    Liu, Xihui},
    journal={arXiv preprint arXiv:2411.07184},
    year={2024}
}

@inproceedings{yang2025omnipart,
  title={Omnipart: Part-aware 3d generation with semantic decoupling and structural cohesion},
  author={Yang, Yunhan and Zhou, Yufan and Guo, Yuan-Chen and Zou, Zi-Xin and Huang, Yukun and Liu, Ying-Tian and Xu, Hao and Liang, Ding and Cao, Yan-Pei and Liu, Xihui},
  booktitle={SIGGRAPH ASIA},
  year={2025}
}

@article{qi2017pointnet,
    author = {Qi, Charles R. and Yi, Li and Su, Hao and Guibas, Leonidas J.},
    title = {PointNet++: deep hierarchical feature learning on point
    sets in a metric space},
    year = {2017},
    journal = {Advances in neural information processing systems}
}

@inproceedings{zhao2021pointtransformer,
    title={Point transformer},
    author={Zhao, Hengshuang and Jiang, Li and Jia, Jiaya and Torr,
    Philip HS and Koltun, Vladlen},
    booktitle=ICCV,
    year={2021}
}

@inproceedings{cen2023segment,
      title={Segment Anything in 3D with NeRFs}, 
      author={Jiazhong Cen and Zanwei Zhou and Jiemin Fang and Chen Yang and Wei Shen and Lingxi Xie and Dongsheng Jiang and Xiaopeng Zhang and Qi Tian},
      booktitle    = NEURIPS,
      year         = {2023},
}

@inproceedings{cen2023saga,
    title={Segment any 3d gaussians},
    author={Cen, Jiazhong and Fang, Jiemin and Yang, Chen and Xie,
    Lingxi and Zhang, Xiaopeng and Shen, Wei and Tian, Qi},
    booktitle=AAAI,
    year={2025}
}

@INPROCEEDINGS{10658260,
    author={Ying, Haiyang and Yin, Yixuan and Zhang, Jinzhi and Wang,
    Fan and Yu, Tao and Huang, Ruqi and Fang, Lu},
    booktitle= CVPR,
    title={OmniSeg3D: Omniversal 3D Segmentation via Hierarchical
    Contrastive Learning},
    year={2024}
}

@inproceedings{gaussian_grouping,
    title={Gaussian Grouping: Segment and Edit Anything in 3D Scenes},
    author={Ye, Mingqiao and Danelljan, Martin and Yu, Fisher and Ke, Lei},
    booktitle=ECCV,
    year={2024}
}

@inproceedings{garfield2024,
    author = {Kim, Chung Min and Wu, Mingxuan and Kerr, Justin* and
    Tancik, Matthew and Goldberg, Ken and Kanazawa, Angjoo},
    title = {GARField: Group Anything with Radiance Fields},
    booktitle = CVPR,
    year = {2024},
}

@article{zhang2024clay,
  title={Clay: A controllable large-scale generative model for creating high-quality 3d assets},
  author={Zhang, Longwen and Wang, Ziyu and Zhang, Qixuan and Qiu, Qiwei and Pang, Anqi and Jiang, Haoran and Yang, Wei and Xu, Lan and Yu, Jingyi},
  journal=TOG,
  year={2024}
}
\end{document}